\documentclass[letterpaper, 10 pt, conference]{ieeeconf}  

\IEEEoverridecommandlockouts                              

\usepackage{algorithm}
\usepackage{algorithmic}
\usepackage{graphicx}
\usepackage{wrapfig}
\usepackage[bookmarks=false]{hyperref}
\usepackage{flushend}
\usepackage{multirow}
\usepackage{caption}
\usepackage{subcaption}
\usepackage{amsmath}
\usepackage{amssymb}

\usepackage{tabularx, booktabs}

\usepackage{mathtools, cuted}
\usepackage{lipsum, color}

\usepackage{mwe,tikz}
\usepackage[percent]{overpic}

\usepackage{tikz}
\usetikzlibrary{shapes,arrows, calc}
\usepgflibrary{arrows}

\usepackage{algorithm}
\usepackage{algorithmic}
\usepackage{graphicx}
\usepackage{wrapfig}
\usepackage[bookmarks=false]{hyperref}
\usepackage{flushend}
\usepackage{multirow}
\usepackage{graphicx}
\usepackage{caption}
\usepackage{subcaption}
\usepackage{amsmath}
\usepackage{amssymb}

\usepackage{txfonts}
\usepackage{mathrsfs}  

\usepackage{mathtools, cuted}
\usepackage{lipsum, color}

\usepackage{tabularx, booktabs}
\usepackage{pbox}
\usepackage{mwe,tikz}
\usepackage[percent]{overpic}
\usepackage{bbm}

\usepackage{tikz}
\usetikzlibrary{shapes,arrows, calc}
\usepgflibrary{arrows}

\DeclareMathOperator*{\argmax}{arg\,max}

\definecolor{arrowblue}{RGB}{98,145,224}

\title{\LARGE \bf
Unsupervised Object Discovery and Segmentation of RGBD-images
}

\author{Johan Ekekrantz$^{*}$, Nils Bore$^{*}$, Rares Ambrus$^{*}$, John Folkesson$^{*}$ and Patric Jensfelt$^{*}$
\thanks{$^{*}$The authors are with the Centre for Autonomous System at KTH Royal Institute of Technology, SE-100 44 Stockholm, Sweden
        {\tt\small \{ekz,nbore,raambrus,johnf,patric\}@csc.kth.se}}%
}

\begin{document}
\maketitle
\thispagestyle{empty}
\pagestyle{empty}

\begin{abstract}
In this paper we introduce a system for unsupervised object discovery and segmentation of RGBD-images. The system models the sensor noise directly from data, allowing accurate segmentation without sensor specific hand tuning of measurement noise models making use of the recently introduced Statistical Inlier Estimation (SIE) method~\cite{SIE}. Through a fully probabilistic formulation, the system is able to apply probabilistic inference, enabling reliable segmentation in previously challenging scenarios. In addition, we introduce new methods for filtering out false positives, significantly improving the signal to noise ratio. We show that the system significantly outperform state-of-the-art in on a challenging real-world dataset.
\end{abstract}

\section{INTRODUCTION}

\begin{figure*}
    \includegraphics[width=0.49\textwidth]{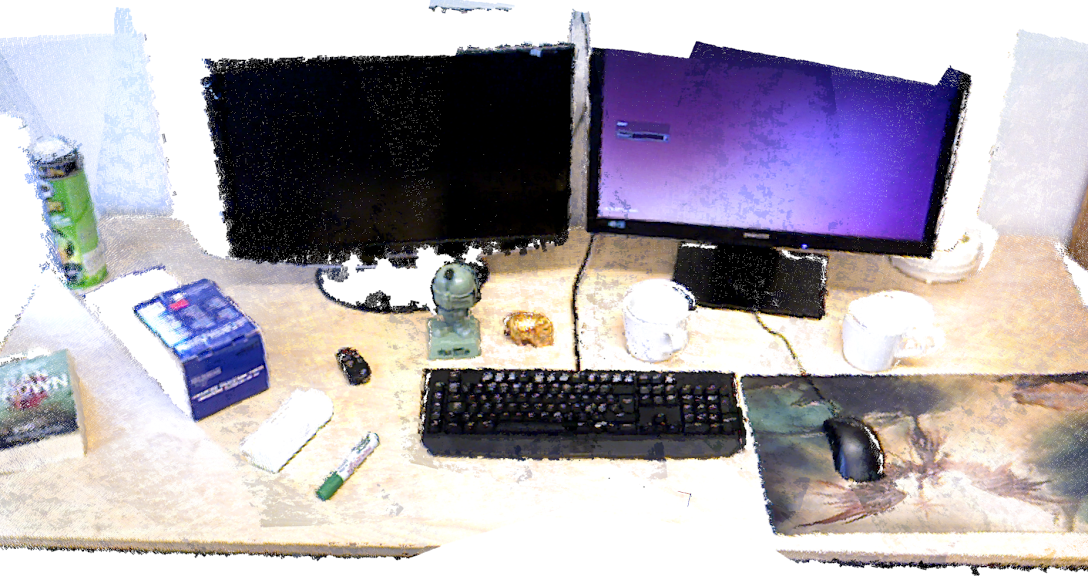}
    \includegraphics[width=0.49\textwidth]{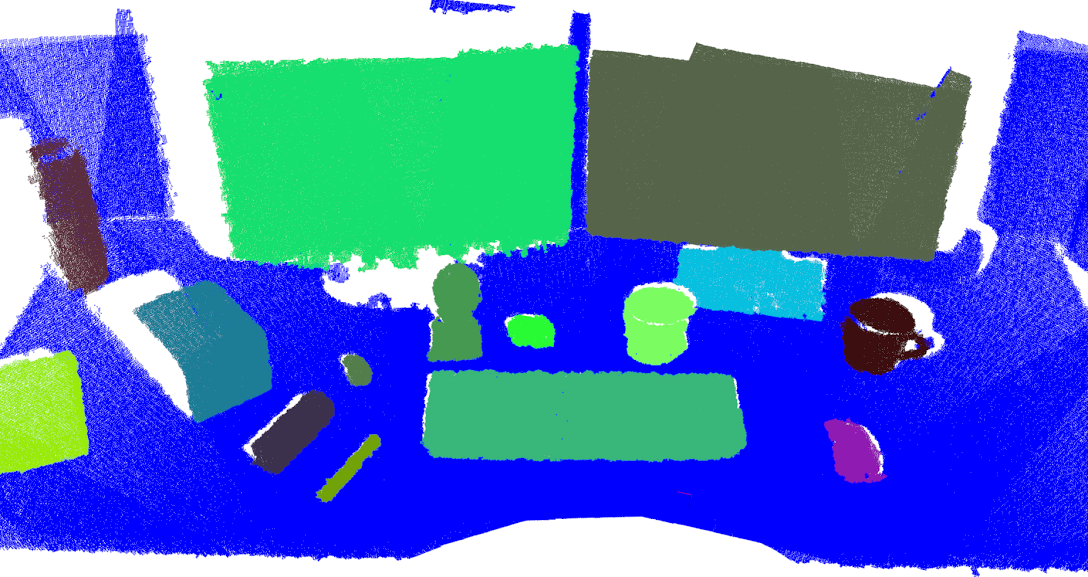}
   \caption{ \\ Left: Sample Desktop scene. Right: Segments found. Blue indicate static background, other objects are randomly colored. The system successfully segment small objects such as cups, a computer mouse, a whiteboard marker and a car key. }
    \label{fig:seg}
\end{figure*}

Object discovery is the task of identifying previously unseen objects. In this paper we do this using RGBD-images.
Segmentation, in the context of this paper, refers to the task of determining which pixels are part of a discovered object. 
Finding and segmenting new objects automatically is challenging because of the diversity of possible shapes and colors of 
both objects and non-objects. In addition, the usual method of finding objects using supervised machine learning techniques is not directly applicable because the task of the system is to discover new objects, as opposed to re-detecting previously seen objects, since there is by definition no information on the new object to train the supervised machine learning algorithm on. Object discovery and segmentation has many potential uses in robotic systems, such as an automatic attention system, anomaly detection or a tool to enable autonomous object learning. 

There are several different methods used to detect objects autonomously, by far the most common solution for RGBD-images rely on the supporting plane assumption, which says that most objects are found on flat supporting surfaces such as desktops, tables or floors.

A second option is to look for repeating structures, since repeating structures in indoor environments often correspond to mass produced objects or functional structures.

Another approach is to perform clustering of superpixels into larger clusters, using for example convexity as a prior for merging superpixels. The clusters can then be classified as either objects or background using some form of scoring criteria.

Change detection is a powerful queue to discover objects. If something new appears in the environment, it is likely to be an object. Given time, everyday objects that people interact with such as chairs, computer mice, keyboards or cups are moved around. Change detection as a methods of detecting objects is very general in that it does not place assumptions on the shape, placement or frequency of objects, allowing a wide variety of objects to be detected. Change detection biases the detections of objects towards finding objects that people interact with on a regular basis and therefore move often, a potentially useful property for human robot interaction as such objects are likely to carry significant meaning or purpose to humans.

Range-sensor such as RGBD-cameras or Lidars measure the distance along some set of directions to the closest surface. The area between a measurement and the sensor is \textit{free space}. A measurement captured at a different point in time which occurs in-between the original measurement and the sensor can either be due to  a new surface appearing or the old surface moving towards the sensor. In literature, such measurements are referred to as occlusions or free space violations. Detection of occlusions form the basis for our change detection since occlusions guarantee change in the underlying 3D structure of the environment. 

In this paper we use introduce a new system for object discovery and segmentation based on change detection through occlusion detection in sets of RGBD-images. A major contribution of the paper is an end-to-end fully probabilistic formulation of the problem. As a means of facilitating a fully probabilistic formulation, a novel method for aggregating data over multiple frames is devised. The method of aggregation does not require a-priori data fusion, but instead perform marginalization over the set of most likely surface configurations given a set of measurements. 

We show that the Statistical Inlier Estimation (SIE) algorithm, which previously has been used for surface shape estimation~\cite{ppr} and pointcloud registration~\cite{SIE}, can be modified and applied to the domains of change detection and probabilistic image edge detection. The SIE algorithm is used to automatically model the sensor noise distribution and perform probabilistic occlusion detection between RGBD-image pairs, removing time consuming sensor and environment specific parameter tuning.

We then apply maximum likelihood inference to infer regions of change between two sets of RGBD-images. As a means of achieving tractable inference over multiple frames simultaneously, lazy optimization is combined with graph-cut energy minimization. Clustering is then applied as a means of finding coherent regions of changed points. 

Areas of change are in previous work~\cite{metarooms}\cite{DBLP:conf/iros/AmbrusBFJ14} called \textit{dynamic} and areas without change are called \textit{static}. In this work we propose to further separate dynamic areas into two subclasses, \textit{moving} for the areas which are currently undergoing change and \textit{moved} for areas which have undergone changes relative to a previous set of observations but are currently static.

We observe that \textit{dynamic} segments which mostly contain measurements of the \textit{moving} type are not likely to contain the type of rigid objects we are interested in modeling. Given an indoor office environment, the primary cause of \textit{moving} segments appear to be people moving about in the environment, there are however many other potential causes such as pets or drapes blowing in the wind. Systematic errors, such as sensor biases and registration failures also appear as \textit{moving}. For the task of object discovery, we can therefore filter out and remove any segments which are primarily \textit{moving} from the set of potential object segments.

Using the probabilistic formulation of this paper, moving objects can be found by performing change detection of a RGBD-image to other RGBD-images captured within a specified time window.

In an indoor office scenario, there are often people sitting still while working. Such people are not completely filtered out by detecting \textit{moving} segments. We therefore apply a pre-trained person detector based on traditional supervised machine learning. 

The supervised person detector and the filtering of \textit{moving} segments complement each other nicely because they fail in very different scenarios. The supervised machine learning person detector fails for partial observations of people as in figure~\ref{fig_p_obj}, while the filtering of \textit{moving} segments fail when people do not move.

\begin{figure*}
    \begin{subfigure}[b]{0.245\textwidth}
        \includegraphics[width=\textwidth]{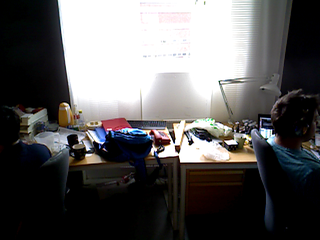}
    \end{subfigure}
    \begin{subfigure}[b]{0.245\textwidth}
        \includegraphics[width=\textwidth]{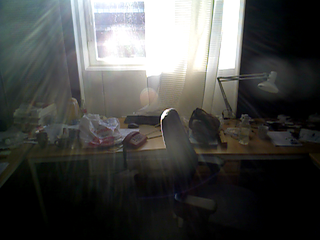}
    \end{subfigure}
    \begin{subfigure}[b]{0.245\textwidth}
        \includegraphics[width=\textwidth]{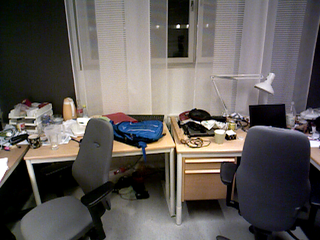}
    \end{subfigure}
    \begin{subfigure}[b]{0.245\textwidth}
        \includegraphics[width=\textwidth]{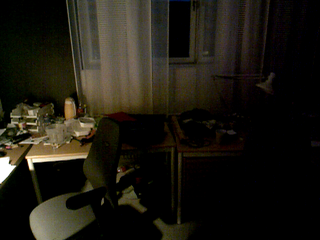}
    \end{subfigure}
    \begin{subfigure}[b]{0.245\textwidth}
        \includegraphics[width=\textwidth]{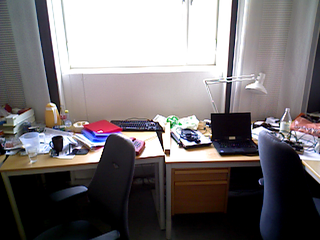}
    \end{subfigure}
    \begin{subfigure}[b]{0.245\textwidth}
        \includegraphics[width=\textwidth]{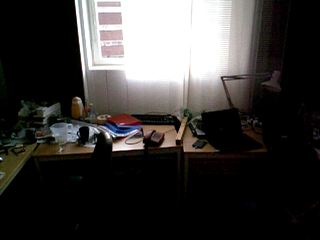}
    \end{subfigure}
    \begin{subfigure}[b]{0.245\textwidth}
        \includegraphics[width=\textwidth]{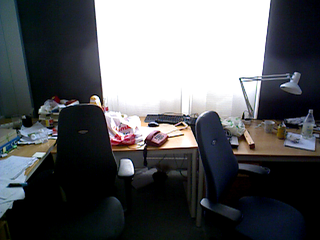}
    \end{subfigure}
    \begin{subfigure}[b]{0.245\textwidth}
        \includegraphics[width=\textwidth]{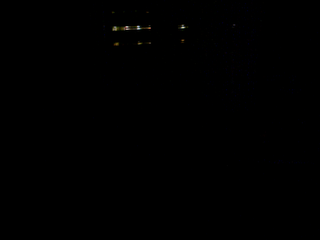}
    \end{subfigure}

   \caption{RGB images of the same desk found in the evaluation dataset. The dataset contains diverse data captured under varying environmental conditions in a cluttered real world environment.}
    \label{fig:datasamples}
\end{figure*}

We evaluate the proposed system on a diverse, realistic and challenging real-world dataset captured autonomously over more than a months time on a mobile robot. In figure~\ref{fig:datasamples} some examples of the  data in the dataset is provided. The proposed system show significant improvements over state-of-the-art on the evaluated dataset. 

An example of the detected and segmented objects found using the proposed system for a desktop scene can be seen in figure~\ref{fig:seg}. As evident from the figure, our system is accurate enough to find even small objects such as car keys and whiteboard markers, using only a cheap commercially available RGBD sensor.

\section{RELATED WORKS}
\label{section:RELATEDWORKS}

Dividing the world into dynamic and static has a long history in both robotics and computer vision.

Background subtraction is a field within computer vision concerned with detect moving objects within a video sequence from a static RGB-camera. A classical application for background subtraction algorithms is video surveillance.
In~\cite{sobral2014comprehensive} a comprehensive review and benchmarking study on the field of background subtraction is presented. Generally background subtraction assumes a static camera and builds models of the background for the pixels in the image plane over many images which are then used to statistically determine if a pixel in a new image contains a moving object or not. In~\cite{fernandez2013background} it was found that incorporating depth information into background subtraction significantly increases the performance of background subtraction in otherwise challenging scenarios. Our scenario differs from background subtraction in that our robot is not static and capture data at unknown intervals. Furthermore, background subtraction is usually performed by comparing the current frame to data aggregated over many previous observations of the same view, whereas our system only require one previous observation of the same scene.

The Simultaneous Localization And Mapping (SLAM) problem has long been central to the performance of mobile robots. Localization using range sensors such as lidars, sonars, radars or structured light sensors is based in the ability to compare current measurements to the predicted state of the environment given previous measurements. Detecting and modelling the static, dynamic and changing areas of an environment is therefore advantageous as shown in~\cite{saarinen2012independent}\cite{kucner2013conditional}\cite{krajnik2014long}. While this work is not focused on the modeling of dynamics in an environment for SLAM, we believe that an accurate detector of static and dynamic areas could prove useful to improve the quality of SLAM solutions.

Automatic object modelling is an application which we predict will be of great importance the the field of robotics once robots become a part of everyday life for most people and an area of research where unsupervised object detection and segmentation plays a key role. In~\cite{ruhnke2009unsupervised}\cite{dimashova2013tabletop}\cite{prankl2015RGB}\cite{valstar14}\cite{marton2010general} plane segmentation is used to find objects placed on flat surfaces such as desktop tables.

Plane detection and segmentation is classically performed using techniques such as RANSAC~\cite{ransac} or the Hough transform~\cite{hough1962method,houghtransform} for pointcloud data but there are also. In~\cite{borrmann20113d} several variations of the Hough transform for plane detection and estimation were evaluated on 3D laser scan data. The Randomized Hough transform~\cite{xu1990new} was found to perform well with regards to computational complexity. Other techniques based on normal clustering~\cite{holz2012real}, the connected component algorithm~\cite{trevor2013efficient}, region growing~\cite{hahnel2003learning}\cite{holz2014approximate} and super pixel merging~\cite{erdogan2012planar} have also been proposed.

In~\cite{ppr} the PPR algorithm was introduced for accurately estimating the parameters and the segmentation of surface primitives such as planes or spheres to pointclouds. The PPR algorithm automatically models the sensor noise and the distribution of outliers. The  PPR algorithm is related to the~\cite{SIE} algorithm used in this paper.

Semantic segmentation is the task of computing  meaningful labels for all pixels in a set of images. Typical labels include wall, floor, furniture, people or object. Our system can therefore be considered to perform semantic segmentation with the classes \textit{moving},~\textit{moved} and~\textit{static}. A popular paradigm for semantic segmentation is to compute pixel priors for each class and pairwise potentials for pixels which are likely to belong to the same class. Statistical inference is then used to infer a maximum likelihood labeling over the pixels. Pixel priors and pairwise potentials are usually found using supervised machine learning techniques. We apply a similar pipeline, however the primary difference to normal semantic segmentation is that we compute pixel priors based on occlusion detection and pairwise priors based on image statistics as opposed to labeled training data. In~\cite{Thoma16a} a review of the field of semantic segmentation is given. In recent years, the field of deep learning have received massive attention due to impressive results, across the board, for supervised machine learning tasks. In~\cite{deepSemanticSegmentationReview} a review on the application of deep learning to semantic segmentation is given.

In this paper we use a standard min-cut/max-flow algorithm~\cite{boykov2004experimental} to perform inference over the conditional random field where pixels are connected in the image plane and on overlapping surfaces in multiple images. The potentials are inferred using the \emph{SIE} method.

In~\cite{krahenbuhl2011efficient} a semantic segmentation approach with an efficiently solved densely connected conditional random field is presented. The approach assumes Gaussian edge potentials and can therefore not be directly applied in our system but shows that densely connected crfs can provide significant advantages over sparsely connected crfs.

In this paper we focus on the detection of objects as surfaces which move. Unsupervised object detection for unorganized collections of RGB images use different queues to detect objects. The first and most common method, co segmentation, detects repeating patterns in images. Finding repeating patterns is suitable for textured objects where finding the same pattern twice is unlikely to be caused by false positive matches. The second method relies on finding sections of the image which visually look like objects. In~\cite{DBLP:journals/corr/HuangRSZKFFWSG016} a CNN based deep learning method was used to learn a bounding box detector for detecting objects in RGB image before image classification is performed. In a sense one can say that the method computes an \textit{objectness} score for a bounding box based on supervised machine learning.

In large datasets, many non objects such as grass or wallpaper texture contain strong repeatability and confuse co segmentation techniques. In~\cite{vicente2011object} co segmentation and the concept of \textit{objectness} is combined to guide the co segmentation towards better object detections.

In~\cite{mattausch2014object} a system for object detection, segmentation and modelling based on planar patches in lidar data is presented.

In~\cite{herbdisc} a system for lifelong object discovery by structured aggregation of multiple different sources of object segmentation information such as objects placed on planar surfaces, \textit{objectness} and repeating appearance.

In~\cite{andreasson2007has} a system for detecting changes in NDT representations based on color images and lidar range scanners is presented.

In~\cite{finman2013toward} a system for automatic object segmentation for RGBD cameras is presented. The system takes as input two maps of an environment in the form of pointclouds. The pointclouds are then compared and the difference computed based on nearest neighbor matching. Segments are then created from clustering on the remaining components and finally filtering is applied to remove false positives. Segments which do not cause sufficient free space violations or are too small are filtered out as false positives. Objects are merged based on feature appearance.

In~\cite{metarooms} a system for autonomous object segmentation for RGBD cameras mounted on a mobile robot is presented. The system uses a pan tilt unit to perform full 360 degree sweeps of an environment. The sweeps are then segmented using point cloud differencing, free space violation detection  filtering and filtering based on cluster size. Objects are merged based on pointcloud registration and spatial modelling of the positions of the object in the environment.

In~\cite{ambrus2015unsupervised} the system of~\cite{metarooms} is improved by including temporal information in the object merging. The improved system uses a big data approach to calibrate the motion of the pan tilt unit, resulting in improved pointcloud alignment and as a consequence significantly improved segmentation. The clusters are filtered to avoid flat segments as such segments are unlikely to be objects. The improved system is evaluated on a dataset recorded autonomously over 30 days using a mobile robot.

In~\cite{conf/icra/HerbstHRF11} a sophisticated measurement model with many parameters is used to perform automatic object segmentation for RGBD cameras based on RGBD motion detection between two RGBD scenes. The measurement model directly compares color values between scenes, which helps detection of changes in the scene if the photometric assumption holds, for example for indoor scenes captured close in time, but is unsuitable when the photometric assumption does not hold, for example long term applications where the compared scenes are captured at different time of day. In~\cite{herbst2011rgb} the approach of~\cite{conf/icra/HerbstHRF11} was extended from pairwise scene comparisons to multiple scene comparisons.

\section{PIPELINE OVERVIEW}
\label{section:OVERVIEW}

\begin{figure*}
	\includegraphics[width=\textwidth]{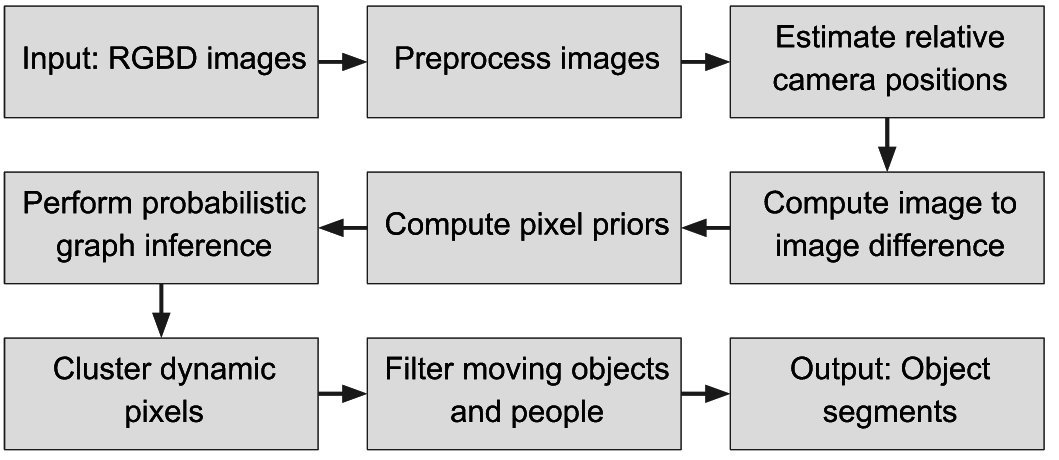}
    \caption{ Object discovery and segmentation pipeline. }
     \label{fig_overview}
\end{figure*}

Our segmentation system takes as input two sets of RGBD-images of a scene, captured at different points in time. By comparing the sets of images, change can be detected and objects segmented. An overview of the object discovery and segmentation pipeline can be found in figure~\ref{fig_overview}.

The first step of the proposed pipeline is to perform pre-processing of the input data to compute surface normals and probabilistic image edges. The edge detector used in this paper uses the SIE algorithm to compute the probability that two neighboring pixels should have the same label. The rest of the algorithm is agnostic to the exact method by which image edges are computed, we therefore place the details on edge detector in the appendix.

After pre-processing the data, the second step of our segmentation pipeline is to estimate the relative positions of all the RGBD-images. Unsurprisingly, the quality of the estimate of the relative positions of the RGBD-images is strongly correlated to the overall performance of the system. Poor estimation of the relative positions reliably lead to static pieces of the environment appearing as dynamic, due to surface self-occlusion. Self occlusion result in false positive detections or under segmentation. As a consequence it is useful to model the uncertainty introduced by imperfect alignment of the input images. The object discovery and segmentation pipeline is independent of the exact system used to estimate the relative camera poses of the input images. The details of the specific solution used align the input data is given in section~\ref{section:EXPERIMENTS}.

We then perform frame differencing and data aggregation to compute the probability of pixels in the input to overlap or occlude previous data or current data. Details on the frame differencing and data aggregation can be found in section~\ref{section:difference}. 

After data aggregation we are able to compute pixel priors for each pixel to be \textit{Object} or \textit{Not Object}. Together with the image edges, we set up a conditional random field to perform probabilistic inference over the labels of all pixels. Details on the probabilistic inference can be found in section~\ref{section:inferencesetup}.

The pixels labeled as \textit{Object} are then clustered to form object hypothesis segments using a variation of the connected component algorithm, see section~\ref{section:clustering} for details. 

The object segments are then classified as either \textit{moving} or \textit{moved} by using the frame differencing from section~\ref{section:difference} to compute the probability that a pixel is currently moving. Objects which are primarily moving are then pruned from the final output as they are unlikely to be true objects. Similarly, any segment which overlap a person as detected by a state-of-the-art deep convolutional neural network person detector is unlikely to be an object and can therefore be pruned from the set of object hypothesises. Details on the filtering of object hypothesises can be found in section~\ref{section:filtering}.

\section{FRAME DIFFERENCING}
\label{section:difference}

In our system, a measurement for a pixel contains a 3D point $\left[ x,y,z \right]^T$, a surface normal $\left[ nx,ny,nz \right]^T$, a color value $\left[ R,G,B \right]^T$ and an estimation $\sigma$ of the noise of the 3D point measurement relative to the other measurements\footnote {We will later use the SIE algorithm to find the absolute scale of the noise. Therefore we only require the relative size of the measurement noise between the 3D-points for now.}. We assume that the measurement noise is zero mean Gaussian and, similarly to~\cite{SIE}, we assume that $\sigma \propto z^2$. We define a function $\delta(I,\{W,H\}) = \big\{ \left[ x,y,z \right]^T, \left[ nx,ny,nz \right]^T, \left[ R,G,B \right]^T, \sigma \big\}$  which extracts a measurement from a RGBD-image $I$ at pixel coordinate $\{W,H\}$.

For a pinhole camera model, we can define the inverse function $\delta^{-1}(p)$ which projects a measurement point $p$ onto the image of the camera, to compute a pixel coordinate correspondence $\{W',H'\}$ as

\begin{equation}
\delta^{-1}(p) = \{ \frac{f_w p_x}{p_z} + o_w,\frac{f_h p_y}{p_z} + o_h \}
\label{eq_reproject}
\end{equation}

Where $o_w$ and $o_h$ is the optical center of the camera used to capture $I$ and the constants $f_w$ and $f_h$ are 
the focal lengths of the camera. Adapting the projection functions to fit other camera models is trivial and therefore left out of this paper.


For each RGBD-image $i$, the relative camera poses is represented as a rigid body transformation consisting of a 3-by-3 rotation matrix $R_i$ and a 3-by-1 translation matrix $t_i$. Using $R_i$ and $t_i$, we define a function $T(i,p)$ which transform a measurement $p$ in the local coordinate system of $i$ to the global coordinate system as

\begin{equation}
\begin{split}
T(i,p) = & \big\{ R_i [p_x,p_y,p_z ]^T + t_i, \\ 
& R_i [p_{nx},p_{ny},p_{nz} ]^T,
 \left[ p_R,p_G,p_B \right]^T, p_\sigma \big\}
\end{split}
\label{eq_transformation}
\end{equation}

and the inverse function $T^{-1}(i,p)$  which transforms a measurement $p$ in the global coordinate system to the local coordinate system of $i$ as

\begin{equation}
\begin{split}
	T^{-1}(i,p) = & \big\{ R^T_i [p_x,p_y,p_z ]^T - R^Tt_i,  \\
	& R^T_i [p_{nx},p_{ny},p_{nz} ]^T,
	 \left[ p_R,p_G,p_B \right]^T, p_\sigma \big\}
\end{split}
\label{eq_invtransformation}
\end{equation}

Given a pair of registered RGBD-images $A,B$ and a measurement $p = \delta(A,\{W,H\})$, for notational convenience defined 
as the input tuple $C = \big \{ A,B,\{W,H\} \big \}$, we would like to determine the probability $P ( O | C )$
that $p$ occlude any measurement in $B$ and the probability $P ( S | C )$, that $p$ is sampled from the same surface patch as a measurement in $B$.
If $P ( O | C )$ is high i.e. $p$ occlude a measurement in $B$, then the measurement $p$ is likely to be part of an object. If on the other 
hand $P ( S | C )$ is high i.e. $p$ comes from the same surface as a point in $B$, $p$ is likely to be part of the static background and not an object. 
If $p$ does not occlude or overlap the data of $B$, there is no direct information about whether $p$ is part of an object or not.

Using eq.~(\ref{eq_transformation}) and eq.~(\ref{eq_invtransformation}), we define the convenience function $\Phi(A,B,\{w,h\})$ that computes the measurement value for pixel $\{w,h\}$ in $A$ and transforms it into the local coordinate space of $B$ as eq.~(\ref{eq_local_to_local}).

\begin{equation}
\Phi(A,B,\{w,h\}) =T^{-1} \bigg( B,T \big( A,\delta(A,\{w,h\}) \big) \bigg)
\label{eq_local_to_local}
\end{equation}

Combining eq.~(\ref{eq_reproject} and eq.~(\ref{eq_local_to_local}), we can define the function $\xi (A,B,\{w,h\})$ that computes the corresponding measurement point in image $B$ for the pixel at $\{w,h\}$ in image $A$ using reprojection as shown in eq.~(\ref{eq_corresponance_measurement}).

\begin{equation}
\xi (A,B,\{w,h\})  = \delta \bigg( B,\delta^{-1} \big( \Phi(A,B,\{w,h\}) \big) \bigg)
\label{eq_corresponance_measurement}
\end{equation}

Given measurement correspondences in a shared coordinate system, direct comparisons can be performed on the measurement values. 
For two points $a$ and $b$, we define the noise normalized point-to-plane distance $\mathcal{D}$ as eq.~(\ref{eq_point_to_plane}.

\begin{equation}
\mathcal{D} (a,b) = \frac{b_{nx}(a_x-b_x)+b_{ny}(a_y-b_y)+b_{nz}(a_z-b_z)}{\sqrt{a_\sigma a_\sigma + b_\sigma b_\sigma}}
\label{eq_point_to_plane}
\end{equation}

Similarly, the noise normalized angular distance $\mathcal{A}$ is computed as eq.~(\ref{eq_angle_distance}) for points $a$ and $b$.

\begin{equation}
\mathcal{A} (a,b) = \frac{1 - a_{nx}b_{nx} - a_{ny}b_{ny} - a_{nz}b_{nz}}{\sqrt{ a_\sigma a_\sigma + b_\sigma b_\sigma}}
\label{eq_angle_distance}
\end{equation}

Combining equations \ref{eq_local_to_local}, \ref{eq_corresponance_measurement}, \ref{eq_point_to_plane} and \ref{eq_angle_distance} allows for straight forward computation of the noise normalized point-to-plane residual $R_\mathcal{D} (C)$ and the noise normalized angular residual $R_\mathcal{A} (C)$ as eq.(\ref{eq_point_to_plane}) and eq.(\ref{eq_angle_distance}) where $C = \big \{ A,B,\{W,H\} \big \}$ is the input tuple, as previously defined.

\begin{equation}
R_\mathcal{D} (C) = \mathcal{D} \big( \xi (C), \Phi (C) \big)
\label{eq_residuals_d}
\end{equation}

\begin{equation}
R_\mathcal{A} (C) = \mathcal{A} \big( \xi(C), \Phi (C) \big)
\label{eq_residuals_a}
\end{equation}

 
Given a set of residuals computed using eq.~(\ref{eq_residuals_d}), we can use the SIE algorithm to determine the probability $P \big( \textrm{overlap} | R_\mathcal{D} (C) \big)$ that $\Phi (C))$ is sampled at the same distance to $B$ as $\xi (C)$. We assume that $R_\mathcal{D} (C)$ follows a zero-mean Gaussian distribution $G ( x,0,\sigma_{R_\mathcal{D} } )$ where SIE is used to estimate $\sigma_{ R_\mathcal{D} }$.

If $\Phi (C))$ and $\xi (C))$ are sampled from the same surface, then the surface normals of $\Phi (C))$ and $\xi (C))$ should also be the same. We then assume that the distribution of residual angular errors from eq.~(\ref{eq_residuals_a}) approximately follows a zero mean generalized Gaussian distribution $\mathcal{G}(x,0,\alpha,\beta) = \frac{\beta}{2 \alpha \Gamma(\frac{1}{\beta})}e^{-0.5*|\frac{x}{\alpha}|^\beta}$, where $\Gamma$ is the gamma function. The Laplacian and Gaussian distributions are special cases of the generalized Gaussian distribution where $\beta = 1$ and $\beta = 2$ respectively. We consider $\alpha$ and $\beta$ to be unknown and use SIE to estimate both. Using the SIE algorithm, we can determine $P \big( \textrm{same normal} | R_\mathcal{A} (C) \big) $ for a set of residuals. Assuming that the measurement noise in $R_\mathcal{D} (C)$ and $R_\mathcal{A} (C)$ is approximately independent, two points are part of the same of the surface if they are overlapping and the normal of both surfaces are similar, then $P ( S | C )$ can be computed as eq.~(\ref{eq_psc}).
 
\begin{equation}
P ( S | C ) \approx P \big( \textrm{overlap} | R_\mathcal{D} (C) \big)  * P \big( \textrm{same normal} | R_\mathcal{A} (C) \big)
\label{eq_psc}
\end{equation}

In theory, one could easily incorporate color by comparing color values as well as surface information in the computation of $P ( S | C ) $. Unfortunately, direct comparisons of color values are sensitive to illumination changes between $A$ and $B$ and therefore not practical without a registration algorithm which also accounts for the nonlinear effects of changing illumination on color values of the RGBD-images. Such effects are small and can potentially be ignored when data is captured within a small difference in time. When data is captured at significantly different points in time, the effects are likely to be severe as seen in figure~\ref{fig:datasamples}.

$P ( O | C ) $ is the probability that $\Phi (C))$ is sampled in-between the image plane of $B$ and $\xi (C)$. Assuming that the measurement noise distribution $G ( x,0,\sigma_{R_\mathcal{D}} )$ for occluding points and the static background is identical, we estimate $P ( O | C )$ by integrating $G ( x,0,\sigma_{R_\mathcal{D}} )$ over $[0,\xi(C)_z]$ and accounting for $ P ( S | C ) $ as eq.~(\ref{eq_poc}).

\begin{equation}
\begin{split}
& P( O | C ) =  P( \neg S | C) \int_{0}^{\xi(C)_z}G( x, R_\mathcal{D} (C), \sigma_{ R_\mathcal{D} })dx
\end{split}
\label{eq_poc}
\end{equation}

The SIE algorithm used to compute $P( S | C )$ and $P( O | C )$ takes as input a set of residuals and outputs an estimate of the measurement noise and a per-residual probability estimate. As usual with estimation algorithms, the quality of estimation is dependent on the amount of data used. 
We therefore aggregate the residuals $R_\mathcal{D}$ and $R_\mathcal{A}$ respectively of all pixels for all frame pairs. 

There are several scenarios where a found correspondence of $\Phi (C))$ and $\xi(C)$ is a priori known to be useless or undefined. We therefore define the function 
$valid(C)$ which indicate if the correspondence is useful and defined. $valid(C)$ is true unless: the re-projection of $\Phi (C))$ ends up outside of the image in $B$, either $\Phi (C))$ or $\xi(C)$ lack depth measurement, $\Phi (C))$ or $\xi(C)$ contain a failed normal estimation. We also consider measurements where $\Phi (C)_z < 0$ invalid, this improves the robustness of the algorithm to inaccurate normal estimation which would in rare cases could cause eq.~(\ref{eq_psc}) to overestimate $P( O | S )$ significantly.

\subsection{Multi-frame aggregation}
\label{section:aggregation}
Our system takes as input a set of previously seen RGBD images to which new images are compared, a method of aggregation of frame-to-frame differences is therefore required. Given that $\textbf{B} = \{ B_0, B_1, \dots , B_m \}$ is the set of previously seen images to which a measurement $p = \delta(A,\{W,H\})$ from image $A$ is compared, for notational convenience defined as the input tuple $\textbf{C} = \big \{ A,\textbf{B},\{W,H\} \big \}$, we would like to determine the probability $P ( O | \textbf{C} )$ that $p$ occlude any measurement in any image in $\textbf{B} $ and the probability $P ( S | \textbf{C} )$, that $p$ is sampled from the same surface patch as a measurement in $\textbf{B}$. 

If all correspondences between $p$ and the images in $\textbf{B}$ were independent, computing $P ( O | \textbf{C} )$ and $P ( S | \textbf{C} )$ would be trivial as they would simply be a product of $P ( O | C )$ and $P ( S | C )$ for all images in $\textbf{B}$. Unfortunately, measurements of a single surface may be present in many of the different images of $\textbf{B}$. The measurements are therefore not independent. Given a surface association for each correspondence, the measurements can be aggregated for each surface individually and independence between the surfaces assured.

As means of facilitating surface data association, we compute the aggregate vector $\textbf{p}^\textbf{B}$ of all valid correspondences to $p$ from the images in $\textbf{B}$, transformed to global coordinates.


If we assume that each correspondence in $\textbf{p}^\textbf{B}$ is associated with a surface which in turn is associated with an index, we can define a vector of vectors $\textbf{Q}$ where $\textbf{Q}_{i,j}$ is the $j$-th image with surface index $i$. For each surface, we aggregate the results by computing the average estimates that $p$ occludes or is sampled from the surface.

Using the now assured independence and aggregation, we compute $ P( S | A , \{w,h\}, \textbf{Q} ) $ as eq.~(\ref{psMerged}) and  $P(O | A , \{w,h\}, \textbf{Q} )$ as eq.~(\ref{poMerged}), where $|\textbf{Q}|$ is the number of of surfaces in $\textbf{Q}$ and $|\textbf{Q}_{i}|$ is the number of images for the $i$-th surface index.

\begin{equation}
P(S | A , \{w,h\}, \textbf{Q} ) = 1 - \prod_{i=0}^{|\textbf{Q}|} \sum_{j=0}^{|\textbf{Q}_i|} \frac{ P( \neg S | A,\textbf{Q}_{i,j},\{w,h\} ) }{ |\textbf{Q}_i| }
\label{poMerged}
\end{equation}

\begin{equation}
P(O | A , \{w,h\}, \textbf{Q} ) = 1 - \prod_{i=0}^{|\textbf{Q}|} \sum_{j=0}^{|\textbf{Q}_i|} \frac{ P( \neg O | A,\textbf{Q}_{i,j},\{w,h\} ) }{ |\textbf{Q}_i| }
\label{psMerged}
\end{equation}

Unfortunately, the correct assignment to $\textbf{Q}$ is unknown and has to be inferred from data. For $\textbf{p}^\textbf{B}_i$ and $\textbf{p}^\textbf{B}_j$, eq.~(\ref{eq_psc}) can be trivially modified to compute the probability that $\textbf{p}^\textbf{B}_i$ and $\textbf{p}^\textbf{B}_j$ are from the same surface as eq.~(\ref{eq_samePoints}).

\begin{equation}
\begin{split}
P ( S | \textbf{p}^\textbf{B}_i , \textbf{p}^\textbf{B}_j) \approx & P \big( \textrm{overlap} | \mathcal{D} (\textbf{p}^\textbf{B}_i ,\textbf{p}^\textbf{B}_j ) \big) \\
\times & P \big( \textrm{same normal} | \mathcal{A} (\textbf{p}^\textbf{B}_i ,\textbf{p}^\textbf{B}_j ) \big)
\end{split}
\label{eq_samePoints}
\end{equation}

For a partitioning $\textbf{K} = \{ K_0,K_1,\dots,K_{n-1} \}$ where $K_i$ is the surface index for $\textbf{p}^\textbf{B}_i$ according to $\textbf{Q}$, the likelihood $\mathcal{L}(\textbf{K},\textbf{p}^\textbf{B})$ of a specific surface partition is the product of all individual pairwise estimates. Going from $\textbf{Q}$ representation of the surface data association to $\textbf{K}$ and back is trivial, for simplicity of notation, we will therefore use the two representations interchangeably.

\begin{equation}
\begin{split}
& \mathcal{L}(\textbf{K},\textbf{p}^\textbf{B}) = \prod_{i=0}^{n}\prod_{j=i+1}^{n} F(i,j,\textbf{p}^\textbf{B}) \\
& F(i,j,\textbf{p}^\textbf{B}) =	
	\begin{cases}
			P ( S | \textbf{p}^\textbf{B}_i, \textbf{p}^\textbf{B}_j) & if K_i = K_j\\
			P ( \neg S | \textbf{p}^\textbf{B}_i, \textbf{p}^\textbf{B}_j) & otherwise\\
	\end{cases}
\end{split}
\label{likelihood}
\end{equation}

If $\mathbb{Q}$ is the complete set of possible surface partitions for $B$, then $P(S | \textbf{C} )$ and $P(O | \textbf{C} )$ can be computed by marginalizing over $\mathscr{Q}$, resulting in eq.(\ref{psWeighted}) and eq.(\ref{poWeighted}) where $|\mathbb{Q}|$ is the total number of surface partitions in $\mathbb{Q}$.

%

\begin{equation}
P(S | \textbf{C} ) = \frac{ \sum_{j=0}^{|\mathbb{Q}|} \mathcal{L}( \mathbb{Q}_j,\textbf{p}^\textbf{B}) P(S | A , \{w,h\}, \mathbb{Q}_j ) }
						  { \sum_{j=0}^{|\mathbb{Q}|} \mathcal{L}( \mathbb{Q}_j,\textbf{p}^\textbf{B}) }
\label{psWeighted}
\end{equation}

\begin{equation}
P(O | \textbf{C} ) = \frac{ \sum_{j=0}^{|\mathbb{Q}|} \mathcal{L}( \mathbb{Q}_j,\textbf{p}^\textbf{B}) P(O | A , \{w,h\}, \mathbb{Q}_j ) }
						  { \sum_{j=0}^{|\mathbb{Q}|} \mathcal{L}( \mathbb{Q}_j,\textbf{p}^\textbf{B}) }
\label{poWeighted}
\end{equation}

The total number of possible partitions of a set, in this case $\textbf{p}^\textbf{B}$, grows at a hyper-exponential rate to the number of of items in the set. It is therefore infeasible to evaluate all possible partitions. Given that the largest effect on the final estimation is had by the maximum likelihood estimate $\textbf{K}_{mle} = \argmax_{\textbf{K}} \mathcal{L}(\textbf{K},\textbf{p}^\textbf{B})$, a solution that ignores the other possible partitions can be a reasonable approximation. An approximate solution to $argmax_{\textbf{K}} \mathcal{L}(\textbf{K},\textbf{p}^\textbf{B})$ can be had through Hierarchical clustering or other greedy data association schemes. While efficient, such solutions does not guarantee that $\textbf{K}_{mle}$ is actually found. We therefore turn to branch-and-bound optimization as a means of computing a set of the most likely possible partitions, this allows us to compute an good approximations of $P(O | \textbf{C} )$ and $P(S | \textbf{C} )$ even when there are multiple likely partitions. We modify the standard branch-and-bound optimization algorithm which finds $\textbf{K}_{mle}$ to instead find all surface partitions $\textbf{K}$ where $ 100 \times \mathcal{L}(\textbf{K},\textbf{p}^\textbf{B}) > \mathcal{L}(\textbf{K}_{mle},\textbf{p}^\textbf{B})$.

Branch-and-bound optimization requires a relaxation scheme to compute a lower-bound estimates. For the optimization to be guaranteed to return the optimal solution, the relaxation has to be optimistic in the terms of the optimization criteria. We define the optimistic relaxation of $F(i,j,\textbf{p}^\textbf{B})$ in eq.(\ref{likelihood}) as $F(i,j,\textbf{p}^\textbf{B}) = max \big( P ( \neg S | \textbf{p}^\textbf{B}_i, \textbf{p}^\textbf{B}_j) , P ( S | \textbf{p}^\textbf{B}_i, \textbf{p}^\textbf{B}_j) \big)$ if either of the variables $i$ or $j$ is to be relaxed.

\section{INFERENCE}
\label{section:inferencesetup}

Objects are made up of coherent surfaces. As a result, objects in images usually contain many connected pixels. 
Similarly, most objects are made up of piece-wise continuously colored surfaces. Therefore, object boundaries in images are usually found where the difference of color values in the image plane is large. Using these assumptions, we would like to segment out coherent objects regions from a set of current images $\textbf{A} = \{ A_0, A_1, \dots , A_n \}$ when compared to a set of background images $\textbf{B} = \{ B_0, B_1, \dots , B_m \}$.

We model the inference of labels of all pixels using a conditional random field (CRF) with the labels $\{ Obj, \neg Obj \}$. Where a pixel with the label $Obj$ indicates that a pixel is part of an object and conversely that a pixel with the label $\neg Obj$ is not part of an object. We maximize the joint probability of the labeling $\textbf{L}$ over all the pixels in all images of $\textbf{A}$ by minimization of a cost function $Cost(\textbf{L}$ as defined in eq.~(\ref{totalCost}), taking into account pixel priors from frame differencing, pixel image plane connectivity and surface overlap between images in $\textbf{A}$. In eq.~(\ref{totalCost}), $\textbf{L}_{i,w,h}$ denote the label for pixel $\{w,h\}$ in image $A_i$.  


\begin{equation}
\begin{split}
& Cost(\textbf{L}) = \sum_{i=0}^{n} \sum_{w,h} \bigg( \overbrace{-Log \big( P(L_{i,w,h} | A_i,\{w,h\}, \textbf{B} ) \big)}^\text{pixel prior} + \\
& \overbrace{CostI(\textbf{L}_i,\textbf{A}_i,w,h)}^\text{image connectivity} + \overbrace{CostS(\textbf{L},\textbf{A},i,w,h)}^\text{surface overlap between images} \bigg)
\end{split}
\label{totalCost}
\end{equation}

\subsection{Pixel Priors}
\label{subsection:priors}

As previously stated, we make the assumption that occlusions are mostly caused by objects. 
We quantify that by specifying $P(Obj | O) = 0.99$. The contribution to $P(Obj | A,\{w,h\},\textbf{B})$ is therefore 
$P(Obj | O) P( O | A , \{w,h\}, \textbf{B} )$. If a measurement does not cause occlusion but 
is sampled from the same surface as previously seen, the probability of that measurement being an object and not a static 
background point is low. We therefore assume that $P(Obj | \neg O , S) = 0.1$. The contribution to $P(Obj | A,\{w,h\},\textbf{B})$ 
is therefore $P(Obj | S) P( \neg O | A , \{w,h\}, \textbf{B}) P( S | A , \{w,h\}, \textbf{B})$. The remaining unknown 
and unaccounted for probability $P(U | A , \{w,h\},\textbf{B} )$ is computed as eq.~(\ref{eq_pu}).

\begin{equation}
\begin{split}
& P(U | A , \{w,h\}, \textbf{B} ) =  P( \neg O | A , \{w,h\}, \textbf{B} )P( \neg S | A , \{w,h\}, \textbf{B})
\end{split}
\label{eq_pu}
\end{equation}


In most applications, false negatives are to be preferred over false positive detections for unknown and poorly connected regions. 
We therefore assume that $P(Obj | U) = 0.49$ which means that the system believes that unknown points are slightly more likely to be non-objects than objects. The contribution to $P(Obj | A,\{w,h\},\textbf{B})$ is $P(Obj | U) P( U | A , \{w,h\}, \textbf{B} )$.

In practice, $P(Obj | O)$ and $P(Obj | \neg O , S)$ are relatively insensitive and result in similar segmentations for a wide range of values, as long as the general assumption that objects move and non-objects are static is maintained.

Finally, $P(Obj | A,\{w,h\}, \textbf{B} )$ can be computed as eq.~(\ref{eq:Pobj}).

\begin{equation}
\begin{split}
& P(Obj | A,\{w,h\}, \textbf{B} ) = \\
& \overbrace{P(Obj | O) P( O | A , \{w,h\}, \textbf{B})}^\text{Occlusion contribution} + \\
& \overbrace{P(Obj | \neg O , S) P( \neg O | A , \{w,h\}, \textbf{B} ) P( S | A , \{w,h\}, \textbf{B} )}^\text{Surface overlap contribution} + \\
& \overbrace{P(Obj | U)P( U | A , \{w,h\}, \textbf{B} )}^\text{Contribution of other cases} \\
\end{split}
\label{eq:Pobj}
\end{equation}

\begin{figure*}
    \begin{subfigure}[b]{0.49\textwidth}
        \includegraphics[width=\textwidth]{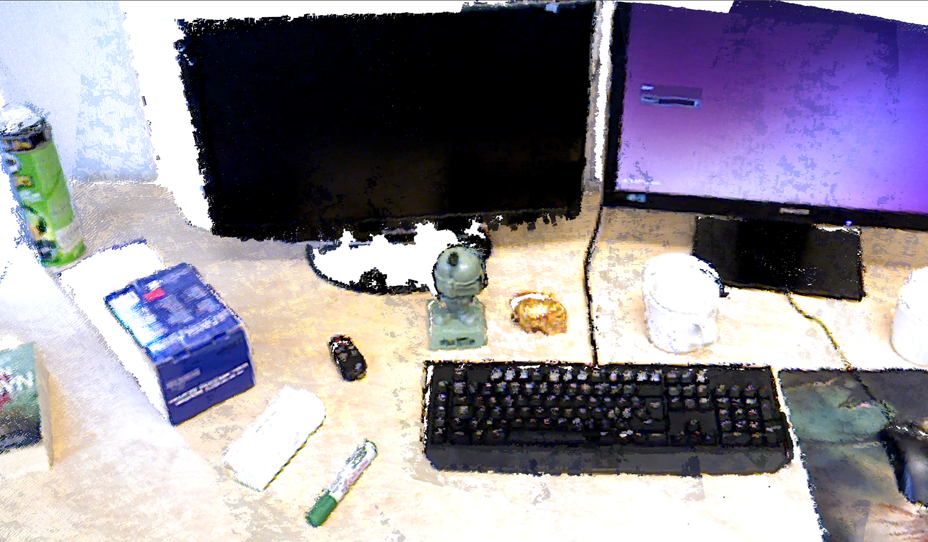}
        \label{subfig_p_obj_rgb}
    \end{subfigure}
    \begin{subfigure}[b]{0.49\textwidth}
        \includegraphics[width=\textwidth]{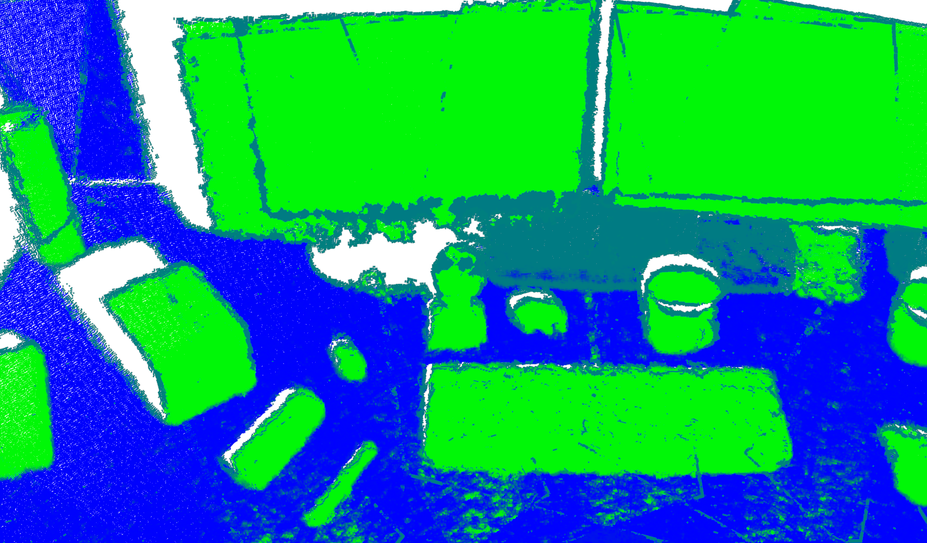}
        \label{subfig_p_obj}
    \end{subfigure}
    \caption{ \\ Left: Colored desktop pointcloud. Right: Estimated probability of being an object shown in green and estimated probability of not being an object shown in blue.}
     \label{fig_p_obj}
\end{figure*}

In figure~\ref{fig_p_obj} we have visualized $P(Obj | A,\{w,h\}, \textbf{B})$ for each point of a desktop scene. The system is able to accurately determine the object boundaries of small objects such as car keys or whiteboard markers thanks to precise noise modelling using the SIE algorithm and careful propagation and aggregation of measurement probabilities.

\subsection{Image Connectivity}
\label{subsection:image_connectivity}

Under the assumption that the real world is made up of piece-wise locally constant colored and connected surfaces, so are the objects we wish to segment. The purpose of the image connectivity cost function $CostI$ is therefore to ensure that neighboring pixels in an image, with similar color and on the same physical surface, have the same label. Assume that $P(p_0 \neq p_1 | A)$ is the probability of pixels $p_0$ and $p_1$ in image $A$ have different color or is part of different surfaces. For convenience, we define a function $\phi(L_0,L_1)$ which takes two labels as input and returns 1 if they are identical and 0 otherwise. $CostI$ can then be computed as eq.~(\ref{eq_costi}) where $p_o = \delta(A,\{w,h\})$ and $L$ is the labels for image $A$.


\begin{equation}
\begin{split}
& CostI(L,A,\{w,h\}) = \\
& -log(\overbrace{ P ( p_o \neq \delta(A,\{w+1,h\}) | A) \big) }^\text{Horizontal connectivity} \phi(L_{w,h},L_{w+1,h})\\
& -log(\overbrace{ P ( p_o \neq \delta(A,\{w,h+1\}) | A) \big) }^\text{Vertical connectivity} \phi(L_{w,h},L_{w,h+1})
\end{split}
\label{eq_costi}
\end{equation}

Many depth sensors have inherent problems with depth values flickering back and forth on or close to depth edges, the depth values close to depth edges are therefore not reliable when computing $P(Obj | A,\{w,h\}, \textbf{B})$. A depth edge is defined as any instance when the probability that the two pixels are sampled from different surfaces is more likely than not. Since the depth data is uncertain around depth edges, we explicitly define $P(Obj | A,\{w,h\}, \textbf{B}) = 0.5$ for such pixels. 

\subsection{Surface Overlap Between Images}
\label{subsection:surface_overlap_between_images}

The purpose of $CostS$ is to ensure that pixels in different images that are sampled from the same underlying surface patch have the same label. The cost for a pixel $\{w,h\}$ in an image $A_i$ is defined as eq.~(\ref{eq_costs}).

\begin{equation}
\begin{split}
& CostS(\textbf{L},\textbf{A},i, \{w,h\}) = \\
& -\sum_{j=0}^{m} \bigg(\phi(L_{i,w,h},L_{j,w',h'})\\
& \times Log \big( \overbrace{ P ( \neg S | A_i,A_j,\{w,h\} ) } ^\text{same surface from frame-to-frame} \big) \bigg)
\end{split}
\label{eq_costs}
\end{equation}

\subsection{Lazy constraint CRF Inference}
\label{subsection:lazyinference}
Given many images to segment, the number of constraints to the CRF from Eq.~(\ref{totalCost}) grows quickly.
Especially the $CostS(\textbf{L},A_i,w,h)$ i.e. the surface overlap between images, grow quickly. Solving the 
full CRF with all constraints can therefore become intractable if there are many overlapping frames. We propose 
to use lazy constraints over $CostS$, a standard technique from the field of discrete optimization, to reduce 
the number of required constraints. Lazy constraint optimization runs as an extension over any discrete 
optimization technique. The optimization is performed without a specific set of constraints (in our case $CostS$). 
Once found, the solution is checked for unused constraints which are violated. Violated constraints are added 
to the optimization problem and the optimization is rerun until convergence. This approach finds the same solution 
as using all constraints, but may only require a fraction of the constraints. We found that for our data, 
between 1 and 3 percent of the constraints in $CostS$ were typically used by the optimization. To avoid adding 
constraints with low influence on the final solution, we use surface overlap constraints where the probability of overlap is greater than 1 percent.

For our application, this makes inference tractable. If, for some other application, the optimization problem becomes 
intractable, applying a discrete optimization meta-heuristic approximate solution such as Large-Neighborhood-Search 
over the segmented objects could significantly speed up the inference.

\section{CLUSTERING}
\label{section:clustering}
From previous sections, probable object pixels have been segmented out from the static background. There are often more than one object in a scene, requiring the object pixels to be clustered into different objects. For this, any suitable clustering technique can be used. \cite{metarooms} uses Euclidean clustering which is the 3D equivalent to the connected component for 2D image data. The Euclidean clustering algorithm has two primary problems, the first is that it requires a hand tuned threshold on the distance between two points under which the points are considered to be connected. This threshold is hard to tune and does not naturally allow to account for the fact that objects far from the sensor are nosier. The second problem with Euclidean clustering is that neighborhood look-up becomes expensive for large pointclouds, making clustering very slow. 

Image plane connected component is on the other hand very fast. Using the previously computed depth edges, we perform image plane connected component using 4-connected pixels. Two neighboring pixels $\{w,h\}$ and $\{w',h'\}$ are considered connected if the probability that they are from the same surface is greater than the probability of being from different surfaces. Similarly we include the surface overlap between images by saying that two pixels in different images $A$ and $B$ are connected if the chance of the surfaces being the same is greater than than the chance of not being the same i.e. $P ( S | A,\{w,h\},B ) > 0.5$ for pixel $\{w,h\}$ in image $A$ and the corresponding pixel in image $B$. The final output of the clustering for a desktop scene is shown in figure~\ref{fig:seg}.

\section{FILTERING}
\label{section:filtering}

As any segmentation approach, our solution occasionally produce imperfect segmentation. Bad segments can be caused by a variety of reasons. The standard approach, which we follow, is to attempt to identify and remove such segments.

The most common form of bad segments are caused by single points or small cluster of points which can be caused by for example jump edges in the depth data or noisy measurements on highly textured areas. For small noisy regions there is simply not enough evidence to support the clusters as reliably being objects. \cite{metarooms} removes such clusters by rejecting clusters where the number of measurements is less than some threshold. 

Poorly calibrated or biased sensors can also lead to bad segments because the distribution of residuals no longer follow the 
assumed measurement model. We also do not consider moving or highly deformable surfaces such as people or pets as objects.
   
We observe that for the set of images $\textbf{A}$, sensor noise, biases and moving surfaces result in self-occlusions. We also
observe that self-occlusion is almost exclusively appears in bad segments in which we are not interested.

We therefore seek to compute a unified metric for the evidence of a cluster being an object, and to not be caused by self-occlusions.

Using the method proposed in section~\ref{section:difference}, we can trivially compute the probability of a pixel being part of a moving object or self occlusion by comparing the current image to other images captured close in time. In practice this means replacing $\textbf{B}$ in $P(Obj | A,\{w,h\}, \textbf{B} )$ from eq.(\ref{eq:Pobj}) with  $ \{x  |  x \subset \textbf{A} \land x \neq A \}$. The probability that a pixel belongs to an object and is not a moving object or self occlusion can therefore be computed as eq.~(\ref{eq_pnotjunk}. The probability that a pixel belongs to an object and is not a moving object or self occlusion can therefore be computed as eq.~(\ref{eq_pnotjunk}). The probability that a pixel is not an object despite occluding previous data is computed as eq.~(\ref{eq_pjunk}).

\begin{equation}
\begin{split}
& P(Junk | A,\{w,h\}, \textbf{B}, \textbf{A}) = \\
& P( Obj | A,\{w,h\}, \{x  |  x \subset \textbf{A} \land x \neq A \} ) P(Obj | A,\{w,h\}, \textbf{B})
\end{split}
\label{eq_pjunk}
\end{equation}

\begin{equation}
\begin{split}
& P(Obj | A,\{w,h\}, \textbf{B}, \textbf{A}) = \\
& P( \neg Obj | A,\{w,h\}, \{x  |  x \subset \textbf{A} \land x \neq A \} ) P(Obj | A,\{w,h\}, \textbf{B})
\end{split}
\label{eq_pnotjunk}
\end{equation}

For each segment $\mathcal{S}$, we compute a score $Score( \mathcal{S} \subset Obj)$ as eq.~(\ref{eq_objscore}) to quantify the number of pixels that are likely to be part of an object and similarly a score  $Score( \mathcal{S} \subset Junk)$ as eq.~(\ref{eq_junkscore}) to quantify the number of pixels that are likely to be some form of false positive.

\begin{equation}
\begin{split}
Score( \mathcal{S} \subset Obj) = & \\
\sum_{A,\{w,h\}} max \big( & P(Obj | A,\{w,h\}, \textbf{B}, \textbf{A}) \\
- & P(Junk | A,\{w,h\}, \textbf{B}, \textbf{A}),0 \big) 
\end{split}
\label{eq_objscore}
\end{equation}

\begin{equation}
\begin{split}
Score( \mathcal{S} \subset Junk) = & \\
\sum_{A,\{w,h\}} max \big( & P(Junk | A,\{w,h\}, \textbf{B}, \textbf{A}) \\
- & P(Obj | A,\{w,h\}, \textbf{B}, \textbf{A}),0 \big)
\end{split}
\label{eq_junkscore}
\end{equation}

We then discard any segment $\mathcal{S}$ which has a higher junk score than object score i.e. $Score( \mathcal{S} \subset Junk) > Score( \mathcal{S} \subset Obj)$. We also discard any segment $\mathcal{S}$ if $Score( \mathcal{S} \subset Obj) < \kappa$ where $\kappa$ controls the minimum amount of support required for a segment to be considered valid and well supported. In our work we set $\kappa = 100$. $\kappa$ is somewhat similar to the threshold used in~\cite{metarooms} to reject of clusters where the number of measurements is low, the main difference is that our formulation account for the uncertainty of the segment as opposed to the size of the segment.



\begin{figure*}
    \begin{subfigure}{0.49\textwidth}
        \includegraphics[width=\textwidth]{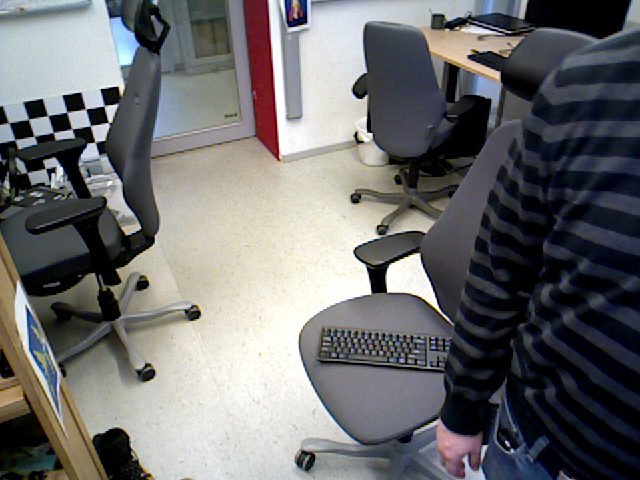}
        \caption{Partial view of moving person captured on camera. Due to partial information, supervised machine learning does not detect the person. \\}
        \label{fig_moving_rgb}
    \end{subfigure}
    \begin{subfigure}{0.49\textwidth}
        \includegraphics[width=\textwidth]{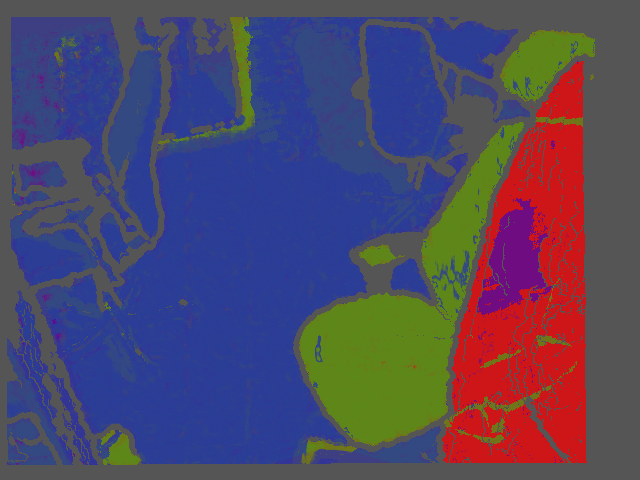}
        \caption{Estimated priors for \textit{moving}, \textit{moved} and \textit{static} in figure ~\ref{fig_moving_rgb} shown in red, green and blue respectively. Notice how the moving person is clearly detected.}
        \label{fig_moving_priors}
    \end{subfigure}
    \\
    \begin{subfigure}{0.49\textwidth}
        \includegraphics[width=\textwidth]{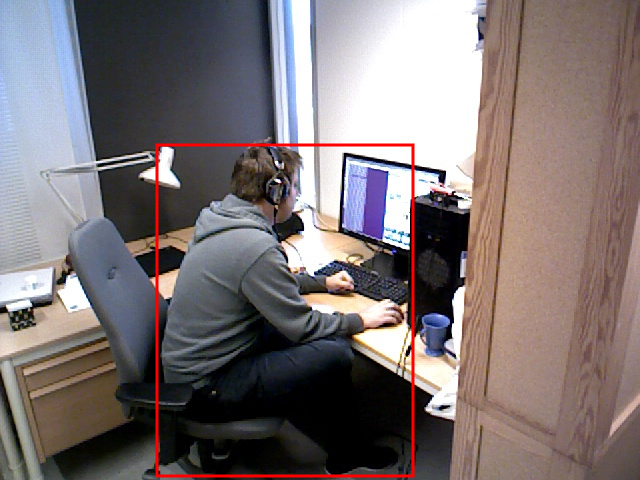}
        \caption{ View of person sitting still and working in-front of a computer. Red rectangle shows person detection by supervised machine learning.}
        \label{fig_static_rgb}
    \end{subfigure}
    \begin{subfigure}{0.49\textwidth}
        \includegraphics[width=\textwidth]{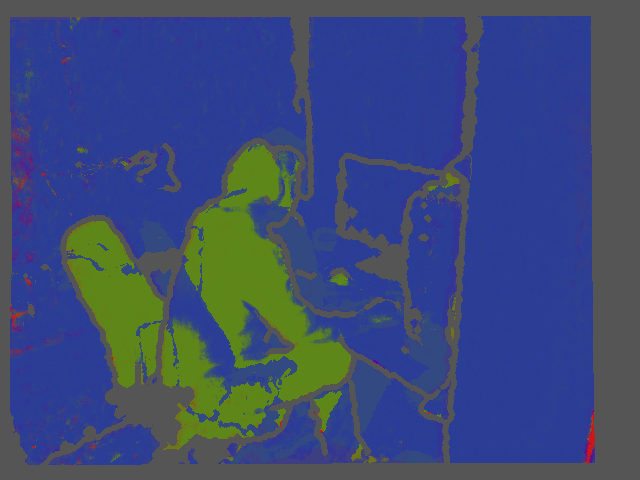}
        \caption{Estimated priors for \textit{moving}, \textit{moved} and \textit{static} in figure ~\ref{fig_static_rgb} shown in red, green and blue respectively. The person sitting still is detected as static and a potential object.}
        \label{fig_static_priors}
    \end{subfigure}
    \caption{Visualization of the complementary of nature of object hypothesis rejection based on supervised machine learning and motion detection.}
    \label{fig_reject}
\end{figure*}


We do not consider people to be objects. We therefore run a pre-trained classifier to separate and remove people from the set of object segments. The classifier creates a bounding box around detected people, objects which overlap such a bounding box are rejected. A similar approach can in theory be applied to other repeatable cases which lead to undesirable segmentations.

The filtering of moving objects and people are complementary in that they can detect different cases where the segmentation fails. The filtering of moving objects fail to detect people which do not move, for example people working in front of computers. The moving object filter on the other hand can remove biases and previously unknown objects, or partial views of moving humans. In figure~\ref{fig_reject} we show the complementary nature of self-occlusion filtering and supervised detections of humans. 


\section{EXPERIMENTS}
\label{section:EXPERIMENTS}

While any two sets of RGBD-images can be used by our system, we use the system of~\cite{metarooms} to autonomously gather data from a Scitos G5 robotic platform. The robot has an Asus Primesense RGBD camera mounted on a pan-tilt-unit, by which a 360 degree sweep of 17 RGBD-images are captured. The robot autonomously patrols around the environment, capturing sweeps at predefined locations in the robots map at semi regular intervals. ~\cite{metarooms} takes advantage of the pan-tilt-unit by pre-computing the relative positions of the different images in the sweeps, resulting in accurate registration of those images at all occasions, regardless of the environmental conditions at the time and location of capture.

As input to our system, we take two sweeps captured at approximately the same location according to the robot localization. The sweeps are then accurately registered using the registration algorithm of~\cite{SIE}.

The experiments will show the performance of our segmentation pipeline, and how we are able to return more accurately segmented objects then previous systems.

\subsection{Dataset}
\label{section:dataset}
To benchmark the proposed solution, we use the dataset of~\cite{metarooms}. The dataset is made up of 360 degree sweeps of a cluttered research lab environment, recorded autonomously on a mobile robot over more than a month of time. The sweeps are performed using an Asus primesense camera on a pan-tilt unit (PTU) on top of a SCITOS G5 mobile robot. A full sweep contain three layers of 360 degree sweeps, each layer containing 17 different RGBD-images. The relative positions of the RGBD-images of sweeps are internally pre calibrated. We limit our experiments the lower layer of each sweep. Our experience is that the vast majority of objects in the scene are found in the lowest layer, as the two topmost layers contain mostly pictures of the ceiling.

We benchmark on sweeps from \textit{WayPoint16}, a busy office environment with many objects. \textit{WayPoint16} contains 92 sweeps and does, as can be seen in figure~\ref{fig:datasamples}, contain cluttered and diverse data with real-world environmental conditions.


\subsection{Experimental Methodology}
\label{section:method}

We compare our solution to the proposed solution of ~\cite{metarooms}, which was designed for the same type of scenario and dataset. The calibration were kept identical for both systems, in order to ensure a fair caparison.
Both solutions were then used to automatically generate object segments, an annotation tool with a graphical user interface was created to annotate the segments based on the accuracy and correctness of the segmentation. Only pixels with valid depth measurements are considered during annotation. The segments created by both algorithms were then annotated by hand and given one out of six labels, as described below:

\begin{description}

\item [\textbf{overlap $\geq$ 0.9}] \hspace{1.3 cm} These segments contain a single object, and the segment masks overlap at least 90 percent of the visible parts of the object. We consider these segments to be of high quality.

\item [\textbf{0.9 $>$ overlap $\geq$ 0.5}] \hspace{2.3 cm} These segments contain a single object, and the segment masks overlap less than 90 percent but overlap at least 50 percent of the visible parts of the object. These segments are usually caused by moderate levels of over segmentation.

\item [\textbf{0.5 $>$ overlap}] \hspace{1.4 cm} These segments contain a single object, and the segment masks overlap less than 50 percent of the visible parts of the object. These segments are usually caused by high levels of over segmentation.

\item [\textbf{Under segmented}] \hspace{2.0 cm} These segments contain a more than one object.

\item [\textbf{Junk}] \hspace{0.1 cm} Any segment that contains pixels which do not correspond to objects are considered to be junk. These segments may contain parts of for example static background structure or humans.

\item [\textbf{Unknown}] \hspace{0.8 cm} Due to the diverse nature of the dataset, the annotator was, in some cases, unable to determine the accuracy or correctness of the segmentation. These segments were found in under or over exposed regions of the images, such as for example dark areas under tables. Such segments were ignored from further analysis.

\end{description}

We investigate the performance of both algorithms with and without human rejection, as human rejection can be easily applied to any object detection and segmentation algorithm to boost the performance, just using simple post-processing of the results.

\subsection{Results and Analysis}
\label{section:results}
We summarize the resulting annotations in table~\ref{results}. We observe that the ratio of junk segments for the proposed solution is significantly lower than the baseline of~\cite{metarooms}. After applying the human rejection, we observe that the ratio of junk segments significantly decrease for both algorithms. For the proposed algorithm, the ratio of junk segments drop to 3.5 $\%$ with human rejection, an order of magnitude lower than the 34 $\%$ of the baseline for~\cite{metarooms}. 

We also observe that the proposed solution finds more than twice the number of segments as compared to ~\cite{metarooms}, both with or without applying human rejection for both algorithms.

The proposed solution also result in a significant increase in high-quality segmentations. With human rejection, 24.5 $\%$ of the total segments found contain high quality segments, 3.6 times higher than the 6.9 $\%$ of~\cite{metarooms}.

Both methods result in a very similar ratio of under segmented objects. We believe that the main cause of under segmentation is having multiple objects moving and placed on-top of close to each other or touching. It is reasonable that both methods would under segment such cases.

\begin{figure*}
    \begin{subfigure}{0.63\textwidth}
        \includegraphics[width=\textwidth]{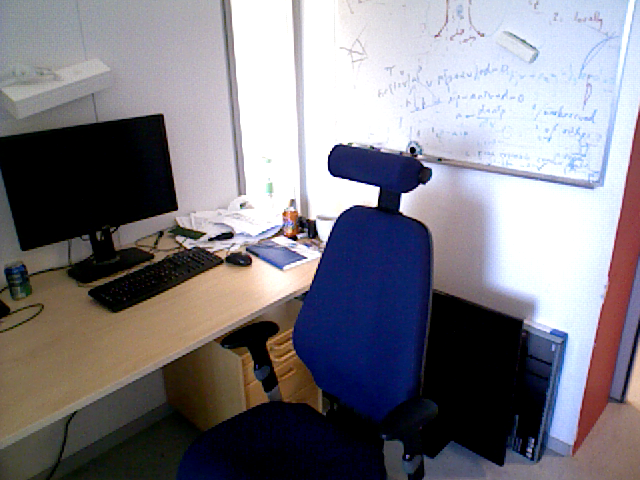}
        \label{fig_oversegmentation_rgb}
    \end{subfigure}
    \begin{subfigure}{0.36\textwidth}
        \includegraphics[width=\textwidth]{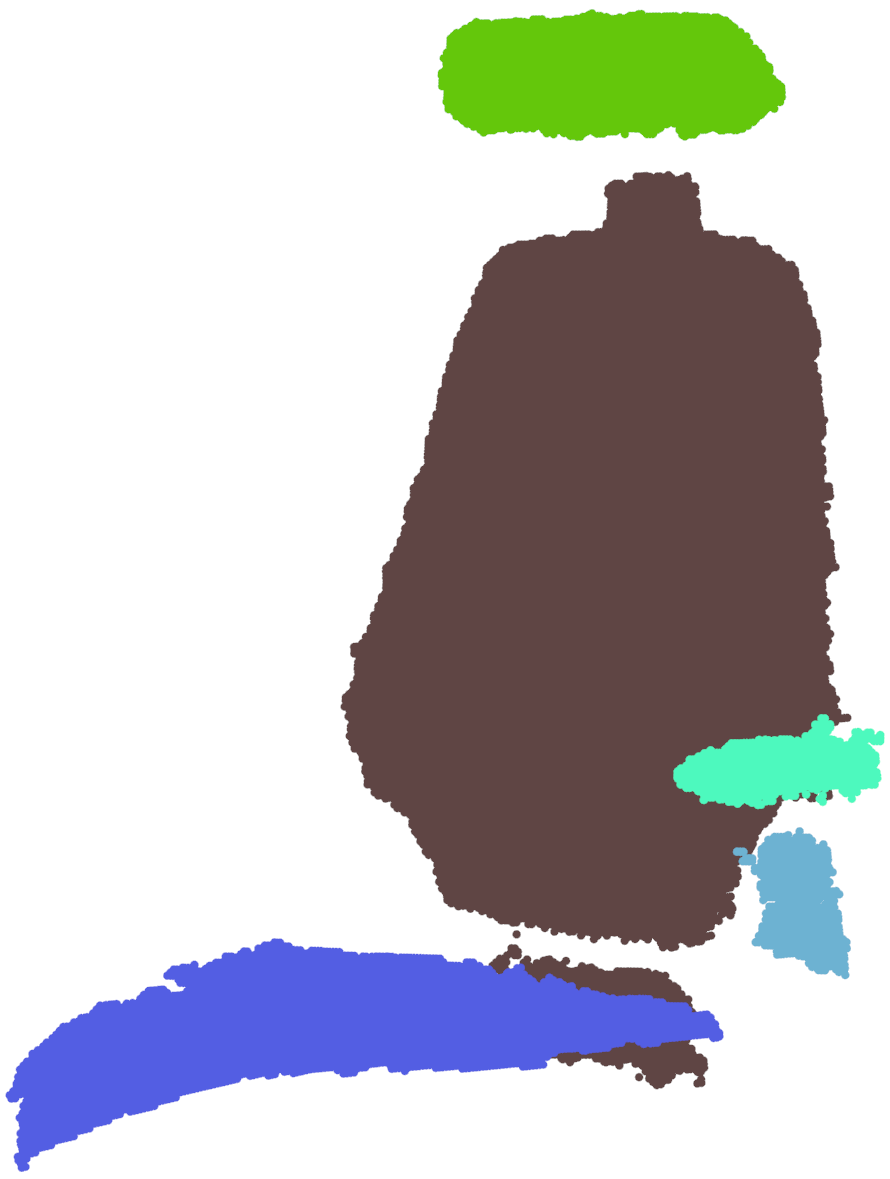}
        \label{fig_oversegmentation_pcd}
    \end{subfigure}
    \caption{Office chairs are complex objects with many self occluding surfaces and metal components which causes depth measurement failures, the connected component or euclidean distance clustering algorithms therefore separate the chair into multiple components. While it could be argued that the segments are correct, due to the ill defined nature of objects, as they make up the neckrest, backrest, seating area and so forth, we consider the chair to be over segmented in this paper.}
    \label{fig_oversegmentation}
\end{figure*}

In the results, we observe that the proposed method outperform the baseline in terms over segmentation, but that both methods leave room for improvement. From direct inspection of the segmentation results, we found that especially office chairs are likely to be made up of multiple disjoint surfaces because of their complex shape and metallic components, leading to over segmentation by both solutions as shown in the example in figure~\ref{fig_oversegmentation}. We believe that data mining techniques looking for repeating constellations of clusters segments could prove a useful took to overcome this type over segmentation. Active exploration where the robot is guided to acquire images from  additional viewpoints to confirm the idea that two segments are indeed from disjoint surfaces would likely reduce the over segmentation as well.


\begin{table*}
\caption{Segmentation experiment results. Ratio of total segments found for each annotation label for our solution and the Metarooms~\cite{metarooms} approach as well as the total number of segments found for both algorithms. }
\centering
    \begin{tabular}{ | l | l | l | l | l | l | l |  l | p{2.5cm} |}
    \hline
    Algorithm 										& overlap $\geq$ 0.9	&  0.9 $>$ overlap $\geq$ 0.5	& 0.5 $>$ overlap	& under segmented		& junk			& total found segments \\ \hline
    \hline
    Metarooms~\cite{metarooms}						& 0.069				& 0.247							& 0.316		  	& 0.031					& 0.337			& 291	\\ \hline
    Metarooms~\cite{metarooms} + people rejection	& 0.	111				& 0.322							& 0.415		  	& 0.047					& 0.105			& 171 	\\ \hline
    Proposed method									& 0.202  			& 0.204							& 0.339	      	& 0.034					& 0.222			& 623	\\ \hline
    Proposed method + people rejection				& 0.249  			& 0.269							& 0.395	      	& 0.050					& 0.035			& 397	\\ \hline

    \end{tabular}
\label{results}
\end{table*}


\section{SUMMARY AND CONCLUSIONS}
\label{section:CONCLUSIONS}
In this paper we presented a system for unsupervised object discovery and segmentation through the use of change and occlusion detection. By adapting and 
extending recent~\cite{SIE} advances from the domain of pointcloud registration to the problem of object discovery and segmentation, we achieve a data-driven solution which adapts
both to the sensor characteristics and environmental conditions without cumbersome hand tuning of sensor noise models. Using a probabilistic formulation, we 
apply statistical inference which enables coherent segmentation over multiple images. By introducing a filter which detects self-occlusion, we can automatically 
detect and filter out moving objects, sensor biases and change detection errors caused by incorrect registration or poor calibration. The system is robust and accurate, 
enabling segmentation of cluttered real world data under various illumination conditions. The segmentation is evaluated on a real-world dataset captured over more 
than a month of time, showing significant improvements over current state-of-the-art. Interesting future work could include tracking segments over time or performing merging of segments based on spatial relations to reduce potential over segmentation.

\section*{ACKNOWLEDGMENT}
This work was funded by SSF through its Centre for Autonomous Systems and the EU~FP7~project~STRANDS~(600623).


\appendix
\label{appendix}

\begin{figure*}
    \begin{subfigure}[b]{0.49\textwidth}
        \includegraphics[width=\textwidth]{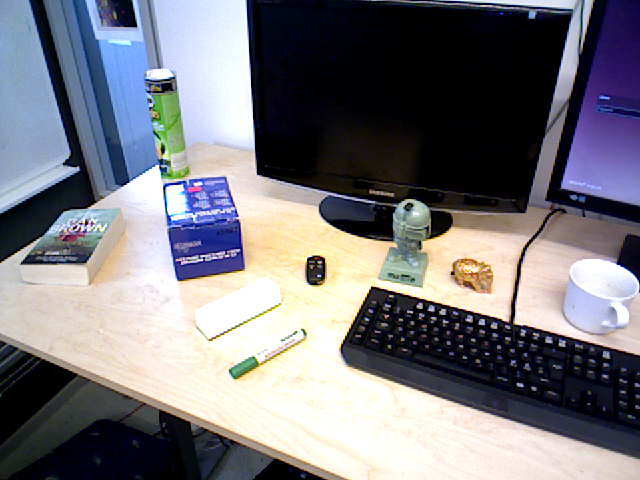}
        \caption{RGB image}
        \label{fig_rgb}
    \end{subfigure}
    \hfill
    \begin{subfigure}[b]{0.49\textwidth}
        \includegraphics[width=\textwidth]{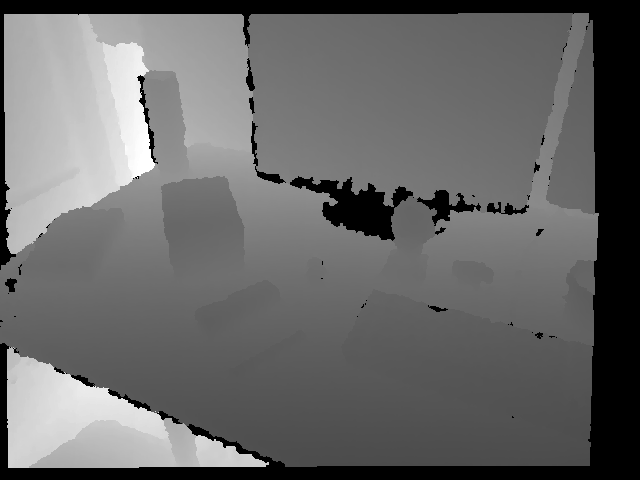}
        \caption{Depth image}
        \label{fig_depth}
    \end{subfigure}
    \begin{subfigure}[b]{0.49\textwidth}
        \includegraphics[width=\textwidth]{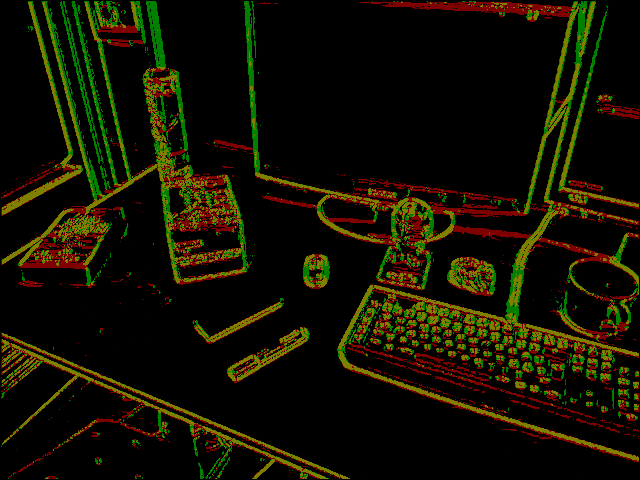}
        \caption{Color edges}
        \label{fig_ce}
    \end{subfigure}
    \hfill
    \begin{subfigure}[b]{0.49\textwidth}
        \includegraphics[width=\textwidth]{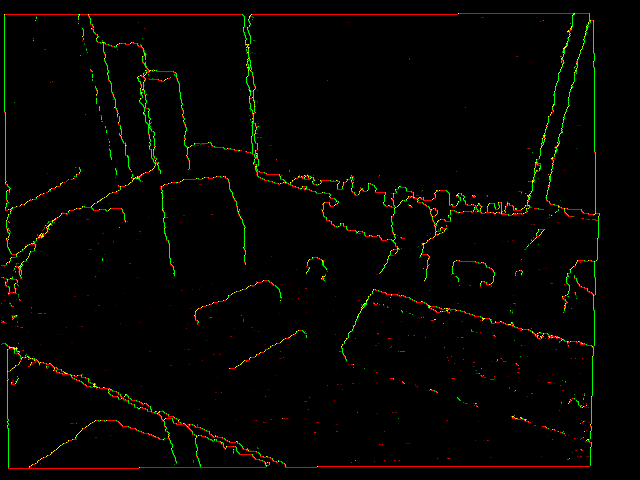}
        \caption{Depth edges}
        \label{fig_de}
    \end{subfigure}
    \caption{RGBD-image and resulting probabilistic edges.}
\end{figure*}

In this paper, we assume that objects are made up of piece-wise continuous surfaces separated by edges in either the color channels or the depth channel in an RGBD-image. An edge in the RGB channels occurs as two neighboring pixels with different colors. Similarly to the seminal work of~\cite{canny}, the first step of finding color image edges is to apply a Gaussian smoothing to the RGB component of the image. The purpose of such smoothing is to reduce the risk of sensor noise to appear as edges. For two neighboring pixels $a$ and $b$ in an image, we define the color residual $R_\mathcal{C}(a,b)$ as eq.(~\ref{eq_rgb_residual}), the three dimensional absolute difference in color values as. We aggregate $R_\mathcal{C}$ for all neighbors in an image. The SIE algorithm is then used to compute the probability $P \big( a_\mathcal{C} \neq b_\mathcal{C} | R_\mathcal{C} (a,b) \big)$ that the color values $a_\mathcal{C}$ from $a$ and the color values $b_\mathcal{C}$ from $b$ have the same underlying color from the area where the pixels are sampled. The advantage of using the SIE algorithm is that the system models the noise on the fly for each image, regardless of image exposure, white balance and other factors which can affect the measurement noise of an image. The SIE algorithm also transform residuals into probabilities which is useful for statistical inference. If the probability that the color values for $a$ and $b$ are different, i.e. $P \big( a_\mathcal{C} \neq b_\mathcal{C} | R_\mathcal{C} (a,b) \big)$ is high, then it is likely that the pixels are also sampled from different surfaces. We denote the probability that $a$ and $b$ are from different surfaces if they are separated by a color edges as $P \big( a \neq b | a_\mathcal{C} \neq b_\mathcal{C} \big)$ which we assume is approximately 0.8. In future work it would be interesting to investigate if $P \big( a \neq b | a_\mathcal{C} \neq b_\mathcal{C} \big)$ can be directly estimated from RGBD-image data on the fly. We can not compute the chance that $a$ and $b$ are from the same surface as~eq.(\ref{eq_p_ab_rgb}).


\begin{equation}
R_\mathcal{C} (a,b) = \bigg\{ | a_R - b_R | , | a_G - b_G | , | a_B - b_B |  \bigg\}
\label{eq_rgb_residual}
\end{equation}

\begin{equation}
\begin{split}
& P \big( a = b | R_\mathcal{C} (a,b) \big) = \\
& 1 - P \big( a \neq b | a_\mathcal{C} \neq b_\mathcal{C} \big) P \big( a_\mathcal{C} \neq b_\mathcal{C} | R_\mathcal{C} (a,b) \big)
\end{split}
\label{eq_p_ab_rgb}
\end{equation} 


We make the assumption that surfaces are locally smooth and that our sensor samples the surface densely. We then define the depth residual between two neighboring pixels as the minimum of the absolute difference in range to the camera for the pixels and the absolute average difference of the pixels as predicted using the local surface slope between the pixels and their neighbors. The residual is then scaled in accordance to the predicted noise of the pixels, using the sum variance law to combine the measurements. Using SIE, we can compute the probability of the pixels being from different surfaces. Similarly to what was done for the color component of the image, the regularization is set to half the discretization step of the sensor. For the our camera the depth discretization step is 0.001 meters. See figures~\ref{fig_depth} and~\ref{fig_de} for an example of a depth image and the found edges. Eq.(\ref{eq_depth_residual}) shows how the depth residual $R$ is computed between pixel $\{w,h\}$ and $\{w+1,h\}$ for RGBD image $I$ where $z_0 = \delta(I,\{w-1,h\})_z$, $z_1 = \delta(I,\{w,h\})_z$, $z_2 = \delta(I,\{w+1,h\})_z$ and $z_3 = \delta(I,\{w+2,h\})_z$. Modifying Eq.(\ref{eq_depth_residual}) to compute the residual between pixel $\{w,h\}$ and $\{w,h+1\}$ is trivial and therefore left out of this paper. 

\begin{equation}
R_\mathcal{Z}(a,b) = \textbf{min} \bigg( \frac{ |z_1-z_2| }{ \sqrt{z_1^4 + z_2^4} } , \frac{ |z_1-2z_2+z_3|+|z_2-2z_1+z_0| }{2\sqrt{z_1^4 + z_2^4} } \bigg) 
\label{eq_depth_residual}
\end{equation}

\begin{equation}
P ( a = b | I) = P \big( a = b | R_\mathcal{Z}(a,b) \big) P \big( a = b | R_\mathcal{C} (a,b) \big)
\label{eq_p_ab}
\end{equation}

Two neighboring pixels are assumed to be part of the same surface if they are not separated by a depth edge and not separated by a color edge. We therefore combine the color and depth as Eq.(\ref{eq_p_ab}) to compute $P ( a = b | I)$, the probability that the neighboring pixels $a$ and $b$ are from the same surface given the image $I$ used to compute the residuals.

\clearpage
\vspace{+60cm}
\newpage
\vspace{+60cm}
\clearpage
\newpage
\vspace{+60cm}


\bibliography{2017_ebafj_segmentation}
\bibliographystyle{ieeetr}

\end{document}